\documentclass{ieeeaccess}
\usepackage{multirow}
\usepackage{cite}
\usepackage{amsmath,amssymb,amsfonts}
\usepackage{algorithmic}
\usepackage{graphicx}
\usepackage{textcomp}
\usepackage{float}
\usepackage{array}
\def\BibTeX{{\rm B\kern-.05em{\sc i\kern-.025em b}\kern-.08em
    T\kern-.1667em\lower.7ex\hbox{E}\kern-.125emX}}
\begin{document}
\history{Date of publication xxxx 00, 0000, date of current version xxxx 00, 0000.}
\doi{10.1109/ACCESS.2017.DOI}

\title{A Time-Frequency based Suspicious Activity Detection for Anti-Money Laundering}
\author{\uppercase{Utku Görkem Ketenc\.{I}}\authorrefmark{1}, 
\uppercase{Tolga Kurt}\authorrefmark{1},
\uppercase{Sel\.{I}m Önal}\authorrefmark{2},
\uppercase{Cenk Erb\.{I}l}\authorrefmark{2}, \\
\uppercase{S\.{I}nan Aktürko\u{g}lu}\authorrefmark{2},
\uppercase{Hande \c{S}erban \.{I}lhan}\authorrefmark{2}}
\address[1]{H3M.IO, ITU TEKNOPARK, ISTANBUL, TURKEY (e-mail: {utku.ketenci, tolga.kurt}@h3m.io)}
\address[2]{AKBANK T.A.Ş, SABANCI CENTER, ISTANBUL, TURKEY (e-mail: {selim.onal, cenk.erbil, sinan.akturkoglu, hande.ilhan}@akbank.com)}

\markboth
{Ketenci \headeretal: Preparation of Papers for IEEE TRANSACTIONS and JOURNALS}
{Ketenci \headeretal: Preparation of Papers for IEEE TRANSACTIONS and JOURNALS}
\corresp{Corresponding author: Utku Görkem Ketenci (e-mail: utku.ketenci@h3m.io).}

\begin{abstract} Money laundering is the crucial mechanism utilized by criminals to inject proceeds of crime to the financial system. The primary responsibility of the detection of suspicious activity related to money laundering is with the financial institutions. Most of the current systems in these institutions are rule-based and ineffective.  The available data science-based anti-money laundering (AML) models in order to replace the existing rule-based systems work on customer relationship management (CRM) features and time characteristics of transaction behaviour. However, there is still a challenge on accuracy and problems around feature engineering due to thousands of possible features.  

Aiming to improve the detection performance of suspicious transaction monitoring systems for AML systems, in this article, we introduce a novel feature set based on time-frequency analysis, that makes use of 2-D representations of financial transactions. Random forest is utilized as a machine learning method, and simulated annealing is adopted for hyperparameter tuning. The designed algorithm is tested on real banking data, proving the efficacy of the results in practically relevant environments. It is shown that the time-frequency characteristics of suspicious and non-suspicious entities differentiate significantly, which would substantially improve the precision of data science-based transaction monitoring systems looking at only time-series transaction and CRM features.

\end{abstract}


\titlepgskip=-15pt

\maketitle

\section{Introduction}
\label{sec:introduction}
Money laundering (ML) is the umbrella under which the legitimization of the proceeds of crime is attempted while laundered money can be both re-inserted into the legitimate economy and re-used to fuel further criminal activities. All major criminality such as drug and human trafficking, terrorism, extortion, kidnap-for-ransom, bribery, embezzlement, tax evasion, corruption and a multiplicity of other offences (also known as predicate offences) are connected through ML. Even though it is impossible to provide an accurate estimate of the size of such a complex underground market, the International Monetary Fund (IMF) indicates that every year, up to 2 trillion USD is laundered through financial systems globally, making ML one of the world's largest markets. 

\begin{figure}[tbp]
\centering
\includegraphics[width=\columnwidth]{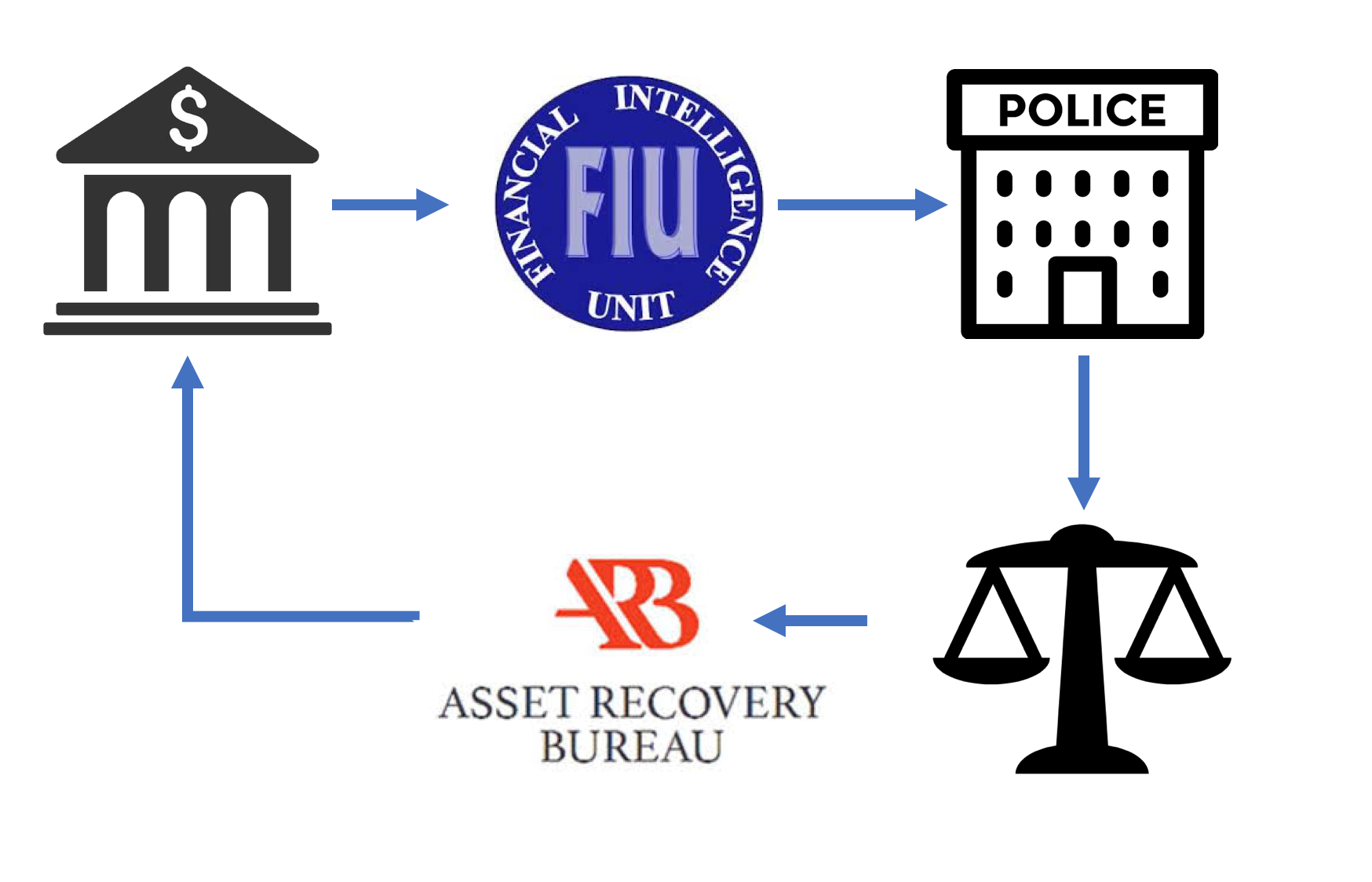}
\caption{The flow of suspicious activity reports through multiple institutions in the judicial system. The detected suspicious activities of bank customers are collected in the Financial Intelligence Unit (FIU), which is reported to the police if necessary. Following the police work, if the judge decides, Asset Recovery Bureau acts upon the recovery process of the laundered money. The process in the jurisdiction of Malta is shown as a representative example.}
\label{figfig}
\end{figure}

To tackle this issue, most countries following the Financial Action Task Force (FATF) recommendations set up an anti-money laundering (AML) structure, as shown in Fig.\ref{figfig}. It is the responsibility of the financial institutions to report suspicious activities to the Financial Intelligence Unit (FIU). The FIU collects intelligence from all different financial institutions within and outside the jurisdiction, which are later reported to the law enforcement agencies (LEA) as necessary. The police, using this intelligence, builds a case to the judicial system, and if ordered the Asset Recovery Bureau, recovers the suspicious assets for public, closing the loop.

As the initiator of the whole process, the identification of the suspicious activity by the financial institutions is very critical. While technology is essential for the processing and identification of suspicious transactions given the volume of data that needs to be filtered, technology adoption in an AML-context needs to be carefully balanced against the various stakeholders in the AML chain of investigation. Also, most of the existing proposed software 'solutions' are rule-based systems with three significant problems. As the first problem, any such software solution depends on a human workforce with varying performance and experience, and instead of enabling AML-analysts and FIU to make more meaningful decisions about what cases should be pursued, they are creating an unmanageable volume of data. With employees being bombarded by a constant stream of noise from technology-based alerts, it is no surprise that negative repercussions are experienced within financial institutions and that these then propagate to FIUs. As the number of false-positive alerts is over 90\% of all alerts, AML experts are consumed by clearing false positives and confirming the non-suspicious nature of technology-generated suspicious cases. This contingency creates difficult work conditions for many AML employees. Often, employees that experience such a continuous stream of false positives will be desensitized towards really suspicious cases.

The second problem is the fact that  most of the suspicious transactions are missed since the rules are exposed to criminals from various channels (e.g. insider threats and employees collaborating with money launderers, reverse engineering of software path-dependencies, published Financial Action Task Force (FATF) typologies that are translated into threshold-based rules or contain specific behavioural traits that can be avoided). As the third problem, the design of rules against new methods of laundering remains a reactive and lengthy process. According to the United Nations Office on Drugs and Crime (UNODC), just 0.2\% of the activities can be detected \cite{united2011estimating}. Despite advances in computation, ML detection remains challenging as a complex behavioural, computational, socio-economic, and managerial problem. 

These problems resulted in the introduction of new methods of transaction monitoring using data science and machine learning techniques. However, most of the machine learning techniques are as successful as the quality of the features that they utilize. There are hundreds of potential features that can be used, such as ATM withdrawals, SWIFT transactions, online transfers, age, occupation. There are also combinations of features that can be created per channel, per time interval, per currency. As a result, feature engineering for AML is a very challenging and time-consuming problem. It can take many weeks, if not months, to determine a useful combination of features out of thousand potential features to be employed.

In this study, we propose a novel and generalized solution using time-frequency (TF) analysis as a feature extraction method, so that with a handful of features, high-level accuracy can be achieved. Time-frequency features improve the accuracy of machine learning results compared to using transaction features alone. The proposed feature set can be utilized as a standard in suspicious transaction detection in order to shorten feature engineering stage. There are three key contributions in this work. The first one is a novel methodology for feature extraction in order to build data science models for AML in time and frequency, significantly reducing feature engineering workload. The second contribution is implementing (in Python) 2D time-frequency features in building data science models for detection, improving the model precision. The third contribution is testing the models in real banking data and proving the improvements in detection of suspicious.

The remainder of the paper proceeds as follows. In the next section, we examine state of the art. In Section III, we present the proposed approach and the time-frequency features. Experiment details and  the experimental results are given in Sections \ref{sec:exp} and \ref{sec:modelperf}, respectively. Finally, we discuss the results and suggest  possible future works.

\section{State of the Art}
\label{sec:state}
\subsection{Data Science Approaches}
The suspicious activity detection rules, in essence, try to model the knowledge of the AML subject matter experts. One of the initial surveys of the application of data mining to AML was given in \cite{watkins2003tracking}. Many approaches work on clustering accounts and transactions and analyzing deviations from clusters and within the cluster for anomalies. For example, Financial Crimes Enforcement Network (FinCEN) has created the FinCEN AI system (FAIS) that links and evaluates reports of large cash transactions to identify potential money laundering; this has been in operation at FinCEN since 1993. The objective is to detect previously unknown, potentially high-value entities (transactions, subjects, accounts) for possible investigation \cite{senator1995financial}.

In one of the first study combining domain knowledge in anti-money laundering and data mining, Zdanowicz \cite{zdanowicz2004detecting} proposed an approach for outlier detection in under- and over-invoicing. However, global outliers correspond only to a small part of money laundering activities. Therefore, more specific approaches have been presented in order to detect local outliers. In \cite{zhu2006outlier}, outliers have been detected with peer group analysis techniques.

\cite{gao2009application,larik2011clustering,liu2011research} applied a cluster-based approach consisting of unsupervised learning techniques such as k-means \cite{macqueen1967some}. Chen et al. \cite{chen2014exploration} improved existing outlier detection results with expectation-maximization technique \cite{dempster1977maximum}. Unlike the general approach to cluster the customers, Soltani et al. \cite{soltani2016new} cluster the transactions and detect money laundering activities according to structural similarity. In \cite{li2009discovering}, the authors propose a new financial transaction grouping method using hidden Markov model and genetic algorithm.

Another well-known approach utilized in the field of AML is supervised machine learning. \cite{tang2005developing} applied support vector machines to suspicious activity detection. Radial basis function neural network has been used in \cite{lv2008rbf}. On the other hand, the decision tree approach has been applied in \cite{wang2007money,ju2009research,villalobos2017statistical,jayasree2017money}. In another recent work \cite{jamshidi2019novel}, adaptive neuro-fuzzy inference system is adopted for the AML problem.

Agent-based approach is proposed in \cite{kingdon2004ai} for the detection of suspicious activity by heterogeneous agents called sentinels. In \cite{wang2007agent}, an agent-oriented ontology for monitoring and detecting money laundering process has been presented. In the same way, a multi-agent system architecture has been examined for AML problem in \cite{liu2007agent}.  In \cite{alexandre2018multi}, the existing multi-agent systems have been extended to combat both fraud and money laundering, and some remarkable results have been presented.

FAIS system had tried data science methods and machine learning, however, due to low ratio of the number of money laundering samples to all financial transactions, there has been a problem of insufficient labelled data for the machines to learn and train. In order to overcome this, active learning has been proposed \cite{deng2009active}. Active learning (AL) is about artificial intelligence asking questions. In order to train the machine better and faster, AL identifies samples that require labelling by AML experts so that future machine-based decisions can be re-oriented based on information that is more reflective of the domain and the ontological nature of suspicion.

\subsection{Time-Frequency Analysis}

The afore-mentioned methods utilize two types of features in data science models; Customer Relationship Management (CRM) features and transaction features in time-domain. In this study, we apply a time-frequency spectrogram analysis of transaction data for suspicious activity detection related to anti-money-laundering.

Time-frequency analysis is one of the most potent tools for time-series analysis and has a wide range of applications in multiple domains from security to image processing. The advantages usually arise from the capability of dividing the signals to numerous components in time and frequency for additional signal processing flexibility. Many techniques have been studied in the last years typically differentiate in the way of changing the signal from time to frequency domain \cite{cohen1995time}. Some of the well-known methods that are utilized are: Fourier transform (FT) consists in decomposing a function into its constituent frequencies, Wigner-Ville distribution, empirical mode decomposition, Gabor transform, Wavelet transform.

To the best of authors' knowledge, time-frequency analysis has never been utilized in the context of AML. The idea presented in this work is to transform transaction data into a 2-dimensional time-frequency representation and use statistical features of the time-frequency domain, instead of features of just transactions in the time domain. Time-frequency analysis can be a powerful tool for AML for the following reasons: First, it can provide a complete picture of characteristics of a single entity and divergence from the peer group; second, it can be representative of behaviour changes from a different perspective. It is expected that the characteristics of routine transactions are much smoother in the time-frequency domain (respectively, suspicious transactions are much sharper) compared to the time domain or frequency domain investigated alone. Thus, it is also expected that time-frequency domain features will be discriminative for the detection of suspicious transactions.

\section{System Model}

\begin{figure}[tbp]
\centering
\includegraphics[width=\columnwidth]{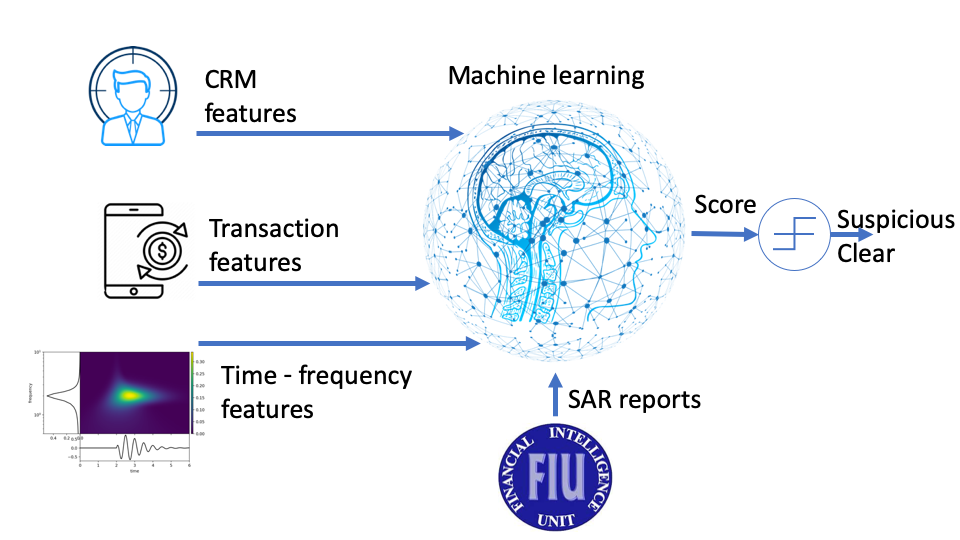}
\caption{System model diagram: The machine learning model (random forest) uses three types of features; CRM features, transaction features, and time-frequency features. The model is trained using previously known suspicious activity. The result of the model is a score between 0 and 1, which is converted into decision by an optimized threshold. The threshold reflects the risk tolerance of the financial institution. }
\label{fig0}
\end{figure}

A simplified block diagram of the proposed system model is presented in Figure \ref{fig0}. The CRM and  transaction database are used for feature creation. Target variable (suspiciousness label) is generated from previously reported suspicious activities. These data sources compose the training set. Machine learning algorithm takes training set as input and create score for all cases. According to a specified threshold, suspicious or clear decision is undertaken. The novelty in the proposed approach is the usage of time-frequency analysis and time-frequency features will be detailed in the following sections.

\begin{figure*}[tbp]
\centering
\includegraphics[trim=0 150 0 150, clip,width=\textwidth]{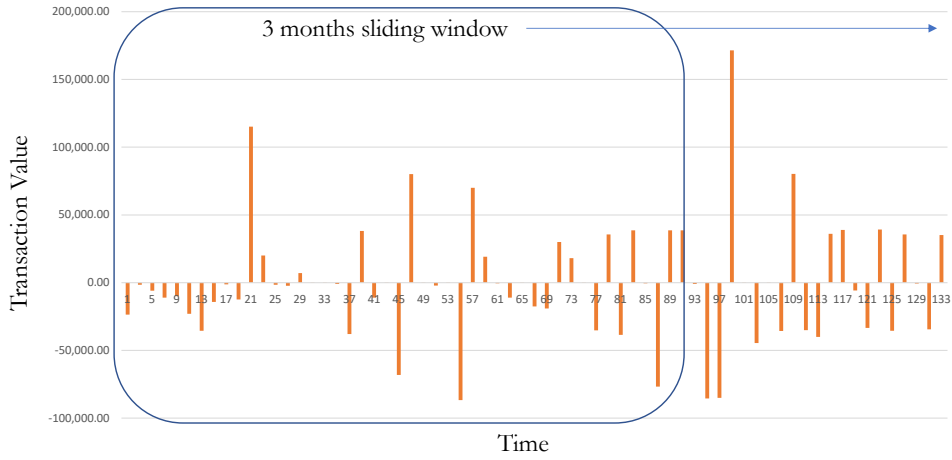}
\caption{Moving time window of transactions (in USD): The bars represent the daily total incoming and outgoing funds, the window of 3 months is slided on daily increments. }
\label{fig1}
\end{figure*}

\subsection{Time-Frequency Representation}

In order to test the effectiveness of the time-frequency features, the following model has been built. For each Akbank customer, the transactions from each banking channel have been modelled as a time series. The funds going in and out to any account of the customer at time $t$ is modelled as $T(t)$. The incoming funds are recorded as positive and outgoing funds as negative.

For a time-series of six months of transactions (signal), $x[n]$, and a time window of $w$, the discrete-time short Fourier transform (STFT) is defined as
\begin{equation*}
STFT(m,w) =  \sum_{-\infty}^{+\infty}x[n] \times w[n-m] \times e^{-j\omega n}. \tag{1}
\end{equation*}

Next, the time-frequency domain representation is formed by moving the three month time window one day at a time for the STFT as in Figure \ref{fig1}. For this study, a quarterly sliding time window is utilized, where the granulization of data is daily, i.e. daily total incoming and ougoing funds form the signal. For futurework, different granulizations and different sizes of time windows as well as their combinations can be utilized.

\begin{figure}[htbp]
\centering
\includegraphics[trim=0 80 0 60, clip,width=\columnwidth]{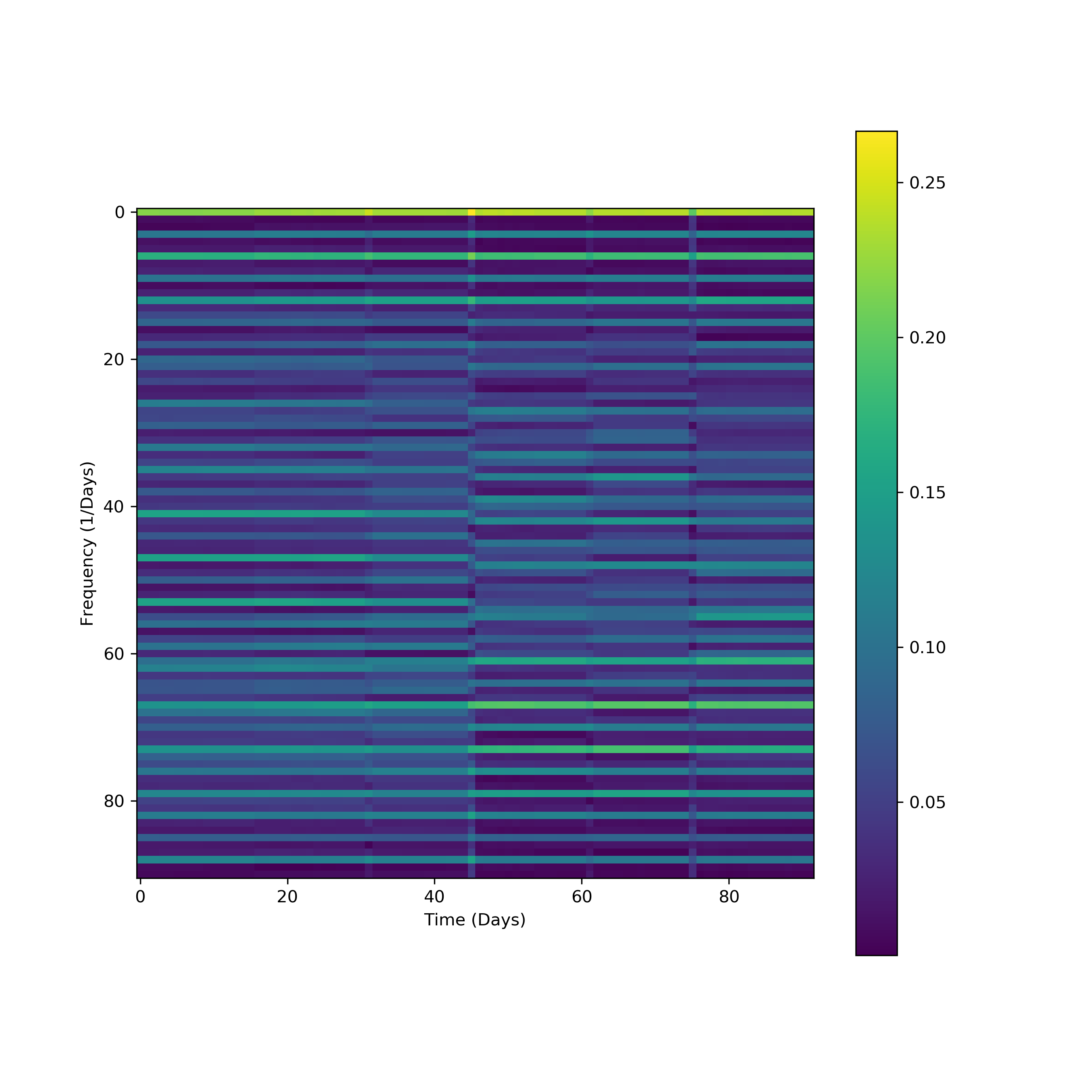}
\caption{Sample normalized time-frequency representation: The time-frequency distribution of a normal customer with expected monthly distributions of fund movements. }
\label{fig2}
\end{figure}

Taking the FT of each window and combining them provides the time-frequency representation of the transactions of a customer, as shown in Figure \ref{fig2}. The figure represents the transactions of a person, who receives salaries on the 15th of every month, pays her rent at the begining of the month, and spends smaller but random amounts on the remaining days. The repetitive and uniform structure of the representation can be clearly seen in the figure.

Feature extraction in this format is in itself a novel approach, which can be utilized by multiple techniques in future studies. In this work, as a first approach, we focus on features that focus on the energy distribution characteristics as explained below.

\subsection{Time Frequency Features}

The idea proposed in this work is that the time-frequency characteristics of customers that use their accounts for everyday financial transactions will be more natural compared to suspicious people. In order to test this idea, the following 11 metrics of the time-frequency domain representation are calculated:
\begin{enumerate}
\item \emph{Mean:} The average value of the time-frequency data points
\item \emph{Variance:} The variance of the time-frequency data points
\item \emph{Skewness:} A measure of asymmetry in the distribution as an investigation of distance from normal distribution
\item \emph{Kurtosis:} The distribution of energy in the FT to the tails of the transformation. In other words, it measures outliers
\item \emph{Time Sparsity:} A measure of sparseness of transactions in the time domain
\item \emph{Frequency Sparsity:} A measure of sparseness of transactions in the frequency domain
\item \emph{Time-Frequency Sparsity:} A measure of sparseness of transactions in the time-frequency domain
\item \emph{Time Discontinuity:} The discontinuity of the frequency distribution as the time-window progress
\item \emph{Frequency Discontinuity:} The discontinuity of the transaction distribution in consecutive frequencies
\item \emph{Time-Frequency Discontinuity:} The discontinuity in the two-dimensional representation
\item \emph{Entropy:} The dispersion of the information content 
\end{enumerate}

\vspace{1\baselineskip}

The hypothesis tested in this work is that, the fundamental energy distribution characteristics of time-frequency representation of suspicious and non-suspicious entities will be different. Using time-frequency features in AML system improvement is the main contribution of this work, whose effectiveness is tested in the next sections.

\section{Experimentation}
\label{sec:exp}
\subsection{Data Set}

Unlike most of the studies in this area, instead of using simulated data, the model is tested with real bank data and actual transactions. In order to build the data science model based on various features, Akbank transaction and CRM data for 6.680 customers are collected, where among 1.945 of them was related to SAR (Suspicious Activity Report) activities and 4.735 of them were not deemed suspicious. 6 months of data were analyzed, in order to have a 90-day iterative shift of 3-month sliding windows of transactions.

\subsection{Model Tuning}

The whole study has been done in a PC having with 16 GB RAM and i7-8700 3.2 GHz CPU.
Random Forest (RF) algorithm \cite{breiman2001random} with 100 trees is adopted for the sake of performance and accuracy issues. RF is a machine learning technique consisting of an ensemble of decision trees. Three parameters of RF are selected for optimization using simulated annealing (i.e. a metaheuristic technique to approximate the global optimum):

\begin{itemize}
\item Minimum numbers of samples to split ($min\_split$): the minimum number of samples required to split an internal node in random forest
\item Minimum number of samples to leaf ($min\_leaf$): the minimum number of samples required to be at a leaf node. A split point at any depth will only be considered if it leaves at least as many training samples as the parameter value in each of the left and right branches.
\item Maximum depth ($max\_depth$): the maximum depth of the tree
\end{itemize}

To investigate the results, the six different cases are considered as features that are input to the data science model:

\begin{enumerate}

\item Training the model with transaction (T) features only
\item Training the model with time-frequency (TF) features only
\item Training the model with customer properties related (CRM) features only
\item Training the model with T + CRM features
\item Training the model with TF + CRM features
\item Training the model with T + TF + CRM
\end{enumerate}

Simulated annealing for 1000 iterations has been run for each feature set for the optimization of min-split, min-leaf and max-depth. Area under the receiver operating characteristic curve (AUC) has been adopted as an objective function to maximize during the optimization.  Optimum parameter sets corresponding to each training set are given in Table \ref{tab:optparam}.

\begin{table}
\caption{The optimum parameters for different training sets of the Random Forest algorithm}
\setlength{\tabcolsep}{3pt}
\begin{tabular}{|p{25pt}|p{60pt}|p{60pt}|p{60pt}|}
\hline
Training Set& 
Optimum $min\_leaf$&
Optimum $min\_split$&
Optimum $max\_depth$
\\
\hline
1&27&30&45 \\
\hline
2&8&17&18\\
\hline
3&18&35&54\\
\hline
4&9&10&45\\
\hline
5&4&12&38\\
\hline
6&8&12&36\\
\hline
\end{tabular}
\label{tab:optparam}
\end{table}

\section{Model Detection Performance}
\label{sec:modelperf}
\subsection{Area Under Curve (AUC)}
\begin{figure}[tbp]
\centering
\includegraphics[width=\columnwidth]{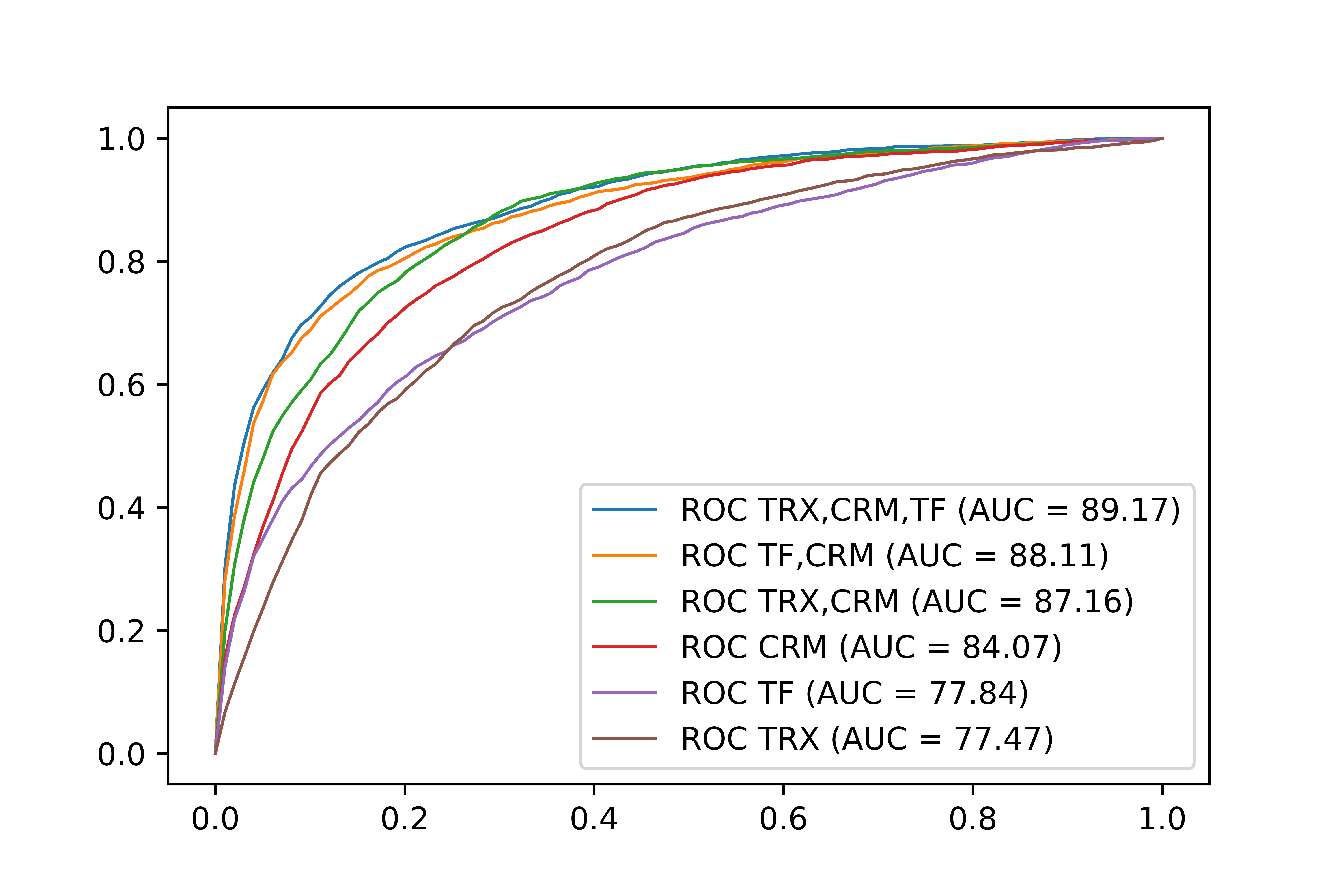}
\caption{ROC curves AUC (in \%) comparisons for different models with respect to the involved features  (TRX: TRANSACTIONS, TF: TIME FREQUENCY, CRM: CUSTOMER PROPERTIES)}
\label{fig3}
\end{figure}

Receiver operating characteristics (ROC) curve represents a true positive rate against false positive rate. AUC corresponds the area under this curve. Higher AUC means higher accuracy. The ROC curves for the six models are shown in Figure~\ref{fig3}.

It can be seen that only time-frequency features and only transaction features provide almost similar performance (respectively 77.47 \% and 77.84 \%). On the other hand, training set with only CRM features give more accurate model with 84.07 \% AUC. At a first glance, we can deduce that traditional CRM features such as age, occupation, etc. are more discriminative.

When time-frequency features are processed with CRM features, the results (88.11 \% AUC) are approximately 1 \% more accurate in terms of AUC comparing to transaction features with CRM features (giving 87.16 \% AUC). Thus, we observe that time-frequency features are more appropriate for combining with CRM features. 

However, we can note that time-frequency and transaction features have complementary effects and AUC is improved additional 1 \% (89.17 \% AUC), once all features are combined. Therefore, AML model becomes more discriminative and more efficient in the detection of suspicious transaction detection.

\begin{table}[tb]
    \begin{minipage}{\columnwidth}
      \caption{Confusion Matrix of model trained model with transaction features only}
      \centering
        \begin{tabular}{|p{25pt}|p{60pt}|p{60pt}|p{60pt}|}
	\hline
	& 
	&
	\multicolumn{2}{c|}{Prediction}
	
	\\
	\cline{3-4}
	& &Negative&Positive\\
	\hline
	\multirow{2}{*}{Reality}&Negative&3350&1385\\
	\cline{2-4}
	&Positive&569&1376\\
	\hline
	\end{tabular}
\label{tab:conf1}
    \end{minipage}%
\vspace{1\baselineskip}
    \begin{minipage}{\columnwidth}
      \centering
        \caption{Confusion Matrix of model trained model with time frequency features only}
        \begin{tabular}{|p{25pt}|p{60pt}|p{60pt}|p{60pt}|}
	\hline
	& 
	&
	\multicolumn{2}{c|}{Prediction}
	
	\\
	\cline{3-4}
	& &Negative&Positive\\
	\hline
	\multirow{2}{*}{Reality}&Negative&3921&814\\
	\cline{2-4}
	&Positive&832&1113\\
	\hline
	\end{tabular}
\label{tab:conf2}
    \end{minipage} 
\vspace{1\baselineskip}

    \begin{minipage}{\columnwidth}
      \centering
        \caption{Confusion Matrix of model trained model with CRM features only}
        \begin{tabular}{|p{25pt}|p{60pt}|p{60pt}|p{60pt}|}
	\hline
	& 
	&
	\multicolumn{2}{c|}{Prediction}
	
	\\
	\cline{3-4}
	& &Negative&Positive\\
	\hline
	\multirow{2}{*}{Reality}&Negative&3594&1141\\
	\cline{2-4}
	&Positive&449&1496\\
	\hline
	\end{tabular}
\label{tab:conf3}
    \end{minipage} 
\vspace{1\baselineskip}
    \begin{minipage}{\columnwidth}
      \centering
        \caption{Confusion Matrix of model trained model with transaction and CRM features}
        \begin{tabular}{|p{25pt}|p{60pt}|p{60pt}|p{60pt}|}
	\hline
	& 
	&
	\multicolumn{2}{c|}{Prediction}
	
	\\
	\cline{3-4}
	& &Negative&Positive\\
	\hline
	\multirow{2}{*}{Reality}&Negative&3827&908\\
	\cline{2-4}
	&Positive&438&1507\\
	\hline
	\end{tabular}
\label{tab:conf4}
    \end{minipage} 
\vspace{1\baselineskip}
    \begin{minipage}{\columnwidth}
      \centering
        \caption{Confusion Matrix of model trained model with time frequency and CRM features}
        \begin{tabular}{|p{25pt}|p{60pt}|p{60pt}|p{60pt}|}
	\hline
	& 
	&
	\multicolumn{2}{c|}{Prediction}
	
	\\
	\cline{3-4}
	& &Negative&Positive\\
	\hline
	\multirow{2}{*}{Reality}&Negative&4218&516\\
	\cline{2-4}
	&Positive&564&1381\\
	\hline
	\end{tabular}
	\label{tab:conf5}
    \end{minipage} 
\vspace{1\baselineskip}
    \begin{minipage}{\columnwidth}
      \centering
        \caption{Confusion Matrix of model trained model with transaction, CRM and time frequency features}
        \begin{tabular}{|p{25pt}|p{60pt}|p{60pt}|p{60pt}|}
	\hline
	& 
	&
	\multicolumn{2}{c|}{Prediction}
	
	\\
	\cline{3-4}
	& &Negative&Positive\\
	\hline
	\multirow{2}{*}{Reality}&Negative&4196&566\\
	\cline{2-4}
	&Positive&507&1438\\
	\hline
	\end{tabular}
	\label{tab:conf6}
    \end{minipage} 
\end{table}

\subsection{0.5 threshold Confusion Matrix}
\label{subsec:confus}
Tables \ref{tab:conf1},\ref{tab:conf2},\ref{tab:conf3},\ref{tab:conf4},\ref{tab:conf5} and \ref{tab:conf6} present 0.5 threshold confusion matrix of model trained respectively with transaction features only, time frequency only, CRM features only, transaction and CRM features, time frequency and CRM features and finally transaction, CRM and time frequency features. In the rows and columns, we present respectively reality (ground truth) and predictions. As aforementioned, predictions are calculated according to 0.5 threshold, i.e. cases having score greater than 0.5 are considered positive (suspicious). The ideal threshold value may vary from institution to institution depending on the risk tolerance and the amount of workload.

We can observe that the number of positive cases predicted as positive (true positive: TP) in Table \ref{tab:conf1} is 263 greater than in \ref{tab:conf2} (and equally the number of negative cases predicted as positive (false positive: FP) in Table \ref{tab:conf1} is 571 greater than in \ref{tab:conf2}). On the other hand,  the number of negative cases predicted as negative (true positive: TN) in Table \ref{tab:conf1} is 571 less than in \ref{tab:conf2} (and equally the number of positive cases predicted as negative (false positive: FN) in Table \ref{tab:conf1} is 263 less than in \ref{tab:conf2}). Thus, we can note that model trained with transaction features are more likely to predict as positive, while the model trained with time frequency features are more likely to predict as negative. 

We observe the same effects between Tables \ref{tab:conf4} and \ref{tab:conf5} combined with CRM features: The number of TP in Table \ref{tab:conf4}  is 126 greater than in \ref{tab:conf5} and the number of TN in Table \ref{tab:conf4} is 391 less than in \ref{tab:conf5}. 

CRM features minimize the number of FN and give the medium number of FP when comparing with transaction and time frequency features results, as we see in Table \ref{tab:conf3}. Finally, best results have been presented in Table \ref{tab:conf6} when combining CRM features with both time frequency and transaction features.

\subsection{False Positive Rate, False Negative Rate and Accuracy}

We calculate the false positive rate (FPR), false negative rate (FNR) and accuracy (ACC) for 0.5 threshold as in (2), (3) and (4). 
\begin{equation*}
FPR = \frac{FP}{FP+TP} \tag{2}
\end{equation*}

\begin{equation*}
FNR = \frac{FN}{FN+TN} \tag{3}
\end{equation*}

\begin{equation*}
Acc = \frac{TP+TN}{TP+TN+FP+FN} \tag{4}
\end{equation*}

The results are presented in Table \ref{tab:fprfnr}. Our findings are summarized below:
\begin{itemize}
\item The best FNR (the minimum value) is achieved with transaction and CRM features. This finding conforms to our assumption (presented in Section \ref{subsec:confus}) saying that transaction features are more likely to predict cases as positive (and less likely to predict cases as negative). Due to low probability to predict negative, there is also low error in negative cases prediction.
\item The best FPR (the minimum value) is achieved with time-frequency and CRM features.  This finding is also in line with our assumption (presented in Section \ref{subsec:confus}) saying that time-frequency features are more likely to predict cases as negative (and less likely to predict cases as positive). Due to low probability to predict positive, there is also low error in positive cases prediction.
\item The best Acc (the highest value) is achieved with transaction, CRM and time frequency. In other words, positive effect of combining transaction and time-frequency features is observed in terms of accuracy for 0.5 threshold.
\end{itemize}

\begin{table}
\caption{False Positive Rate, False Negative Rate and Accuracy comparison according different training sets}
\setlength{\tabcolsep}{3pt}
\begin{tabular}{|p{100pt}|p{35pt}|p{35pt}|p{35pt}|}
\hline
Training Set& 
FPR&
FNR&
Acc
\\
\hline
Transaction Features&29.25\%&29.25\%&70.75\% \\
\hline
Time Frequency Features&17.19\%&42.78\%&75.36\%\\
\hline
CRM Features&24.10\%&23.08\%&76.20\%\\
\hline
Transaction and CRM Features&19.18\%&22.52\%&79.85\%\\
\hline
Time Frequency and CRM Features&10.90\%&29.00\%&83.83\%\\
\hline
Transaction, CRM and Time Frequency Features&11.89\%&26.07\%&84.00\%\\
\hline
\end{tabular}
\label{tab:fprfnr}
\end{table}

\subsection{Importance of Time-frequency features}

Mutual Information (also known as Information Gain) is generally accepted metric for feature importance in data science solution. It's first defined in \cite{shannon2001mathematical}. The mutual information of two jointly continuous random variables $X$ and $Y$ is calculated as 
\begin{equation*}
MI(X;Y) = \int_{Y}^{}\int_{X}^{} p_{X,Y}(x,y)\log\left (  \frac{p_{X,Y}(x,y)}{p_{X}(x)p_{Y}(y)} \right )dxdy \tag{5}
\end{equation*}
where  $p_{X,Y}$ is the joint probability density function of $X$ and $Y$, and $p_{X}$ and $p_{Y}$ are the marginal probability density functions of $X$ and $Y$ respectively. 

In our case, $X$ represents a feature in training set and $Y$ represents its suspiciousness. The relative importance of the features are shown in Table \ref{tab:comp}. More important features have greater mutual information values.

\begin{table}
\caption{Comparison of Mutual Informations of transaction, CRM and time frequency features}
\setlength{\tabcolsep}{3pt}
\begin{tabular}{|p{15pt}|p{140pt}| >{\raggedleft\arraybackslash} p{65pt}|}
\hline
Rank &
Feature & 
Mutual information ($10^{-4}$) \\
\hline
1&AGE&747.35\\ \hline 
2&GENDER&671.17\\ \hline 
3&IS-COMMERCIAL&631.07\\ \hline 
4&RISK-GROUP&613.54\\ \hline 
5&OUTGOING AMOUNT OF FUNDS&612.64\\ \hline 
6&OCCUPATION&610.09\\ \hline 
7&INCOMING AMOUNT OF FUNDS&566.34\\ \hline 
8&CUSTOMER-AGE&341.81\\ \hline 
9&KURT&279.49\\ \hline 
10&SKEW&276.27\\ \hline 
11&ENTROPY&224.76\\ \hline 
12&FDISC&171.43\\ \hline 
13&MEAN&166.06\\ \hline 
14&FSPAR&154.78\\ \hline 
15&FTSPAR&149.48\\ \hline 
16&TDISC&123.29\\ \hline 
17&FTDISC&117.45\\ \hline 
18&TSPAR&76.90\\ \hline 
19&VAR&17.35\\ \hline 
\end{tabular}
\label{tab:comp}
\end{table}

CRM features such as age, gender, boolean showing commercial usage of account, risk group, occupation and number of years of the customer in the bank are among the most important features. We can deduce that CRM features are more discriminative features. On the other hand, these features can be simply evaluated by analysts at first glance and can be more commonly utilized features by analysts. We can also consider these to be the features that most affect analysts' decisions, due to its simplicity.

In Table \ref{tab:comp}, we can observe that transaction features such as incoming and outgoing amount of funds are as important as CRM features. However, when combined with CRM features, the contribution of the transaction features to AUC score (presented in Section \ref{sec:modelperf}) are less than the time frequency features. The explication can be the correlation of incoming and outgoing amount of money between them (approximately 98 \%). Because of the high correlation, these features cannot have an additional effect on accuracy. On the other hand, time frequency features are diverse and not all correlated. Therefore, their usage in training set are more productive in terms of AUC.

The analysis shows that smoothness-based features such as Kurtosis and Skewness are more discriminative among time frequency features. These features are essential in terms of suspicious activity detection, which is in-line with our assumption regarding the suspicious activity characteristics that the time frequency distribution of normal activities are smooth.

\begin{figure*}[tbp]
\centering
\includegraphics[trim=110 200 110 150, clip,width=\textwidth]{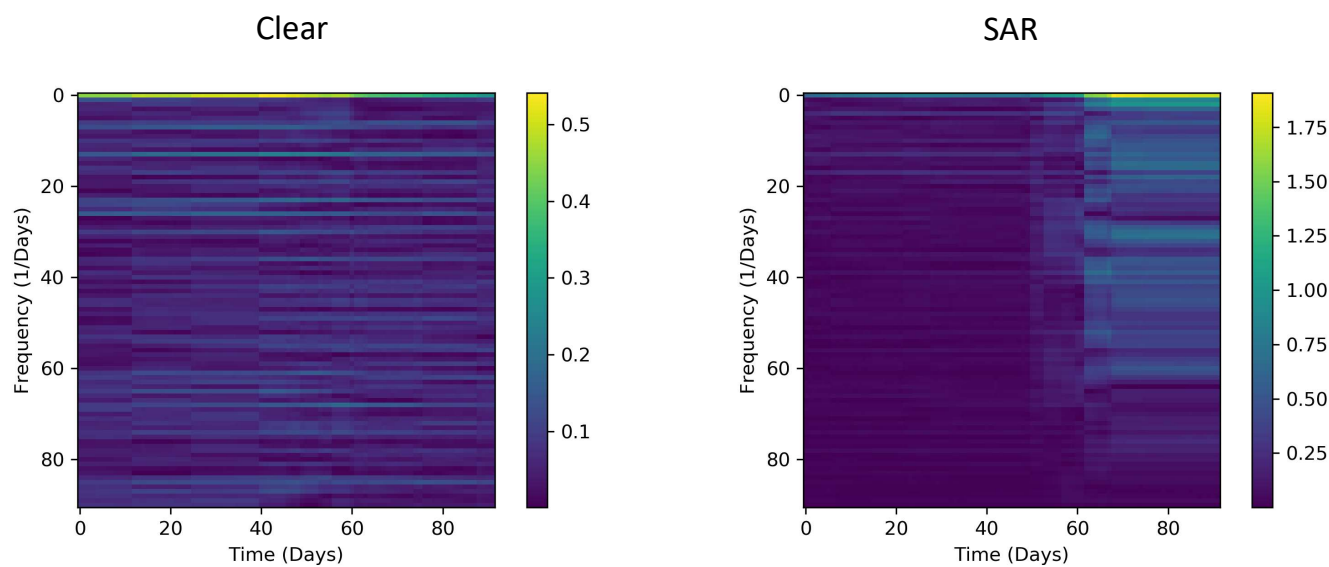}
\caption{Time-frequency plots for sample clear and suspicious cases as a visual for change in account behaviour}
\label{fig4}
\end{figure*}

As an example of suspicious activity, in Figure \ref{fig4}, an account behaviour change is shown, where the transaction frequency distribution changes significantly and sharply. The change can be seen on the right-hand side of the figure; the frequency characteristics differ significantly for an account whose control has been taken by illicit means.

\section{Conclusion and Future work}

In this paper, we have shown that adding time-frequency features, simplifies the feature selection process and improves the quality of the data science model. Time-frequency features such as mean, variance, Kurtosis, skewness have been used for the first time in machine learning model training for suspicious transaction detection. Therefore, feature engineering stage can be shortened by calculating the proposed time-frequency feature set. This potentially saves many person-months of modelling studies for the financial institutions. 

The utility of our approach has been implement in Python and demonstrated with high-level accuracy achieved by testing on real financial data. The generalized and simplified solution can easily be adapted to detect suspicious transactions in various organization. An analysis of actual customer data indicates that time-frequency features can distinguish between suspicious and clear cases, improving AUC and the efficiency of the transaction monitoring system. From different time-frequency characteristics, Kurtosis provided the maximum differentiation in the model. The gains in accuracy and detecting money laundering cases that were not detectable before, can save financial institutions form regularity fines in the order of millions of USD.

In this work, only a low complexity Fourier transform-based approach is utilized for frequency domain analysis. As a future work, the time-frequency analysis can be accomplished with other types of linear and non-linear transforms. There are also potential gains in comparing multiple window lengths, increment sizes and making the analysis in multiple dimensions (such as ATM, Branch, Web). Also, the same analysis can be extended to investigate the characteristics of networks rather than single entities. In particular, when a customer has multiple accounts in multiple banks, the whole picture can only be analyzed by the FIUs. Therefore, repeating this study in FIU data would be beneficial as well. Hence, time-frequency features have numerous potential future uses in the area of financial behaviour analysis.

\bibliography{mybib}{}

\begin{thebibliography}{10}

\bibitem{united2011estimating}
U.~N.~O. on~Drugs and Crime, ``Estimating illicit financial flows resulting
  from drug trafficking and other transnational organized crimes,'' 2011.

\bibitem{watkins2003tracking}
R.~C. Watkins, K.~M. Reynolds, R.~Demara, M.~Georgiopoulos, A.~Gonzalez, and
  R.~Eaglin, ``Tracking dirty proceeds: exploring data mining technologies as
  tools to investigate money laundering,'' {\em Police Practice and Research},
  vol.~4, no.~2, pp.~163--178, 2003.

\bibitem{senator1995financial}
T.~E. Senator, H.~G. Goldberg, J.~Wooton, M.~A. Cottini, A.~U. Khan, C.~D.
  Klinger, W.~M. Llamas, M.~P. Marrone, and R.~W. Wong, ``Financial crimes
  enforcement network {AI} system ({FAIS}) identifying potential money
  laundering from reports of large cash transactions,'' {\em AI magazine},
  vol.~16, no.~4, pp.~21--21, 1995.

\bibitem{zdanowicz2004detecting}
J.~S. Zdanowicz, ``Detecting money laundering and terrorist financing via data
  mining,'' {\em Communications of the ACM}, vol.~47, no.~5, pp.~53--55, 2004.

\bibitem{zhu2006outlier}
T.~Zhu, ``An outlier detection model based on cross datasets comparison for
  financial surveillance,'' in {\em IEEE Asia-Pacific Conference on Services
  Computing (APSCC)}, pp.~601--604, 2006.

\bibitem{gao2009application}
Z.~Gao, ``Application of cluster-based local outlier factor algorithm in
  anti-money laundering,'' in {\em IEEE International Conference on Management
  and Service Science}, pp.~1--4, 2009.

\bibitem{larik2011clustering}
A.~S. Larik and S.~Haider, ``Clustering based anomalous transaction
  reporting,'' {\em Procedia Computer Science}, vol.~3, pp.~606--610, 2011.

\bibitem{liu2011research}
R.~Liu, X.-l. Qian, S.~Mao, and S.-z. Zhu, ``Research on anti-money laundering
  based on core decision tree algorithm,'' in {\em IEEE Chinese Control and
  Decision Conference (CCDC)}, pp.~4322--4325, 2011.

\bibitem{macqueen1967some}
J.~MacQueen {\em et~al.}, ``Some methods for classification and analysis of
  multivariate observations,'' in {\em Proceedings of the Fifth Berkeley
  Symposium on Mathematical Statistics and Probability}, vol.~1, pp.~281--297,
  Oakland, CA, USA, 1967.

\bibitem{chen2014exploration}
Z.~Chen, A.~Nazir, E.~N. Teoh, E.~K. Karupiah, {\em et~al.}, ``Exploration of
  the effectiveness of expectation maximization algorithm for suspicious
  transaction detection in anti-money laundering,'' in {\em IEEE Conference on
  Open Systems (ICOS)}, pp.~145--149, 2014.

\bibitem{dempster1977maximum}
A.~P. Dempster, N.~M. Laird, and D.~B. Rubin, ``Maximum likelihood from
  incomplete data via the {EM} algorithm,'' {\em Journal of the Royal
  Statistical Society: Series B (Methodological)}, vol.~39, no.~1, pp.~1--22,
  1977.

\bibitem{soltani2016new}
R.~Soltani, U.~T. Nguyen, Y.~Yang, M.~Faghani, A.~Yagoub, and A.~An, ``A new
  algorithm for money laundering detection based on structural similarity,'' in
  {\em IEEE Annual Ubiquitous Computing, Electronics \& Mobile Communication
  Conference (UEMCON)}, pp.~1--7, 2016.

\bibitem{li2009discovering}
Y.~Li, D.~Duan, G.~Hu, and Z.~Lu, ``Discovering hidden group in financial
  transaction network using hidden {Markov} model and genetic algorithm,'' in
  {\em IEEE International Conference on Fuzzy Systems and Knowledge Discovery},
  vol.~5, pp.~253--258, 2009.

\bibitem{tang2005developing}
J.~Tang and J.~Yin, ``Developing an intelligent data discriminating system of
  anti-money laundering based on svm,'' in {\em IEEE International conference
  on machine learning and cybernetics}, vol.~6, pp.~3453--3457, 2005.

\bibitem{lv2008rbf}
L.-T. Lv, N.~Ji, and J.-L. Zhang, ``A rbf neural network model for anti-money
  laundering,'' in {\em IEEE International Conference on Wavelet Analysis and
  Pattern Recognition}, vol.~1, pp.~209--215, 2008.

\bibitem{wang2007money}
S.-N. Wang and J.-G. Yang, ``A money laundering risk evaluation method based on
  decision tree,'' in {\em IEEE International Conference on Machine Learning
  and Cybernetics}, vol.~1, pp.~283--286, 2007.

\bibitem{ju2009research}
C.~Ju and L.~Zheng, ``Research on suspicious financial transactions recognition
  based on privacy-preserving of classification algorithm,'' in {\em IEEE First
  International Workshop on Education Technology and Computer Science}, vol.~2,
  pp.~525--528, 2009.

\bibitem{villalobos2017statistical}
M.~A. Villalobos and E.~Silva, ``A statistical and machine learning model to
  detect money laundering: an application,'' 2017.

\bibitem{jayasree2017money}
V.~Jayasree and R.~S. Balan, ``Money laundering regulatory risk evaluation
  using bitmap index-based decision tree,'' {\em Journal of the Association of
  Arab Universities for Basic and Applied Sciences}, vol.~23, pp.~96--102,
  2017.

\bibitem{jamshidi2019novel}
M.~B. Jamshidi, M.~Gorjiankhanzad, A.~Lalbakhsh, and S.~Roshani, ``A novel
  multiobjective approach for detecting money laundering with a neuro-fuzzy
  technique,'' in {\em IEEE International Conference on Networking, Sensing and
  Control (ICNSC)}, pp.~454--458, 2019.

\bibitem{kingdon2004ai}
J.~Kingdon, ``Ai fights money laundering,'' {\em IEEE Intelligent Systems},
  vol.~19, no.~3, pp.~87--89, 2004.

\bibitem{wang2007agent}
Y.~Wang, D.~Xu, H.~Wang, K.~Ye, and S.~Gao, ``Agent-oriented ontology for
  monitoring and detecting money laundering process,'' in {\em Proceedings of
  the International Conference on Scalable Information Systems}, pp.~1--4,
  2007.

\bibitem{liu2007agent}
X.~Liu and P.~Zhang, ``An agent based anti-money laundering system architecture
  for financial supervision,'' in {\em IEEE International Conference on
  Wireless Communications, Networking and Mobile Computing}, pp.~5472--5475,
  2007.

\bibitem{alexandre2018multi}
C.~Alexandre and J.~Balsa, ``A multi-agent system based approach to fight
  financial fraud: An application to money laundering,'' 2018.

\bibitem{deng2009active}
X.~Deng, V.~R. Joseph, A.~Sudjianto, and C.~J. Wu, ``Active learning through
  sequential design, with applications to detection of money laundering,'' {\em
  Journal of the American Statistical Association}, vol.~104, no.~487,
  pp.~969--981, 2009.

\bibitem{cohen1995time}
L.~Cohen, {\em Time-frequency analysis}, vol.~778.
\newblock Prentice Hall, 1995.

\bibitem{breiman2001random}
L.~Breiman, ``Random forests,'' {\em Machine learning}, vol.~45, no.~1,
  pp.~5--32, 2001.

\bibitem{shannon2001mathematical}
C.~E. Shannon, ``A mathematical theory of communication,'' {\em The Bell System
  Technical Journal}, vol.~27, no.~3, pp.~379--423, 1948.

\end{thebibliography}
\bibliographystyle{ieeetr}

\clearpage

\begin{IEEEbiography}[{\includegraphics[width=1in,height=1.25in,clip,keepaspectratio]{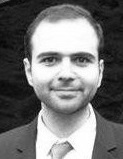}}]{Utku Görkem Ketenci} received the B.S. degree in Computer Science from the Galatasaray University, Istanbul, in 2008; the M.S. degree in Computer Science from University Joseph Fourier, Grenoble, in 2009 and the Ph.D. degree in Computer Science from University of Valenciennes, Valenciennes, in 2013.

From 2013 to 2014, he was a Research Assistant with the DIM (Décision, Interaction, Mobilité) team of LAMIH from University Polytechnique of Valenciennes.
Since 2014, he is working as Data Scientist in Banking and Finance Sector. He is the author of several articles in leading journals and conference proceedings.
His research interests include multi-agent systems and machine learning.
\end{IEEEbiography}

\begin{IEEEbiography}[{\includegraphics[width=1in,height=1.25in,clip,keepaspectratio]{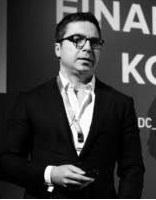}}]{Tolga Kurt} received his B.S. and M.S. degrees from Bo\u{g}azi\c{c}i University, Turkey in Electrical and Electronics Engineering in 2000 and 2002 respectively, and  Ph.D. in Electrical Engineering from the University of Ottawa, Canada, in 2006. Following his Ph.D., he was responsible for the 4G product line in Ericsson Canada. Dr. Kurt later managed his first startup PlusOneMinusOne working in operational optimisation for Banks. 

Currently, he is the managing partner of AI-based AML solutions company H3M.IO. Due to his work in big data analytics, he has been selected to MIT Technology Review 35 under 35 list and have been selected as the Young Entrepreneur of the Year in Turkey. He has extensive R\&D project management experience, which has been translated into productization in more than 40 countries. He has managed more than 15 international software projects, 25 national R\&D projects supported by Tubitak and 2 EU projects. Dr. Kurt has published more than 5 patents, 10 journal papers and 30 international conference papers. 
\end{IEEEbiography}

\begin{IEEEbiography}[{\includegraphics[width=1in,height=1.25in,clip,keepaspectratio]{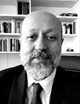}}]{Sel\.{I}m Önal} received the Political Sciences degree in Istanbul University, Istanbul. From 1991 to 1999, he worked as an Internal Auditor at T.C. Ziraat Bank. From 1999 to 2004 he worked as a Risk Manager at Retail Banking at Koçbank (currently Yapı Kredi Bank). Since 2005, he is working as a Chief Compliance Officer at Akbank Compliance Department.

He is responsible for Akbank Group Compliance management which covers all affiliates in Turkey, Germany and Malta. He participates and leads Turkish Banking Association (TBA) Compliance Working Groups such as TBA-MASAK (Turkish FIU) Working Group, National Risk Asssesment Group and Digital KYC Group. He attended as Turkish Banking Sector delegate to FATF and Private sector meetings in Paris, Brussels and Amsterdam. He also participated as a speaker in many international conferences in London, Paris, Amsterdam, Brussels, Berlin, Istanbul, Ankara and Antalya.

\end{IEEEbiography}
\begin{IEEEbiography}[{\includegraphics[width=1in,height=1.25in,clip,keepaspectratio]{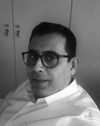}}]{Cenk Erb\.{I}l} started his banking career in 2000 at Koçbank, after graduating from Marmara University, Business Administrations Department in German. For 15 years of his 20 years banking experience, he has been working at the Compliance Department of Akbank T.A.Ş. He is working as a Financial Crimes Compliance Vice President at Akbank Compliance Department.
\end{IEEEbiography}
\begin{IEEEbiography}[{\includegraphics[width=1in,height=1.25in,clip,keepaspectratio]{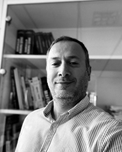}}]{S\.{I}nan Aktürko\u{g}lu} received the Faculty of Management in Kocaeli University, in 2002 and received Management MBA in Kadir Has University in 2008. From 2002 to 2004,  he worked as an Operation Leader at Martaş Port Companies. From 2004-2007, he worked as a Marketing Manager at family business operating in the textile industry.  Since 2007, he has been working at the Compliance Department of Akbank T.A.Ş. He is working as a Financial Crimes Investigation and Compliance Manager at Akbank Compliance Department.
\end{IEEEbiography}
\begin{IEEEbiography}[{\includegraphics[width=1in,height=1.25in,clip,keepaspectratio]{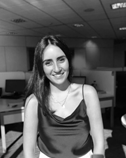}}]{Hande \c{S}erban \.{I}lhan} received the International Relations degree in İstanbul University, İstanbul in 2014. From 2015 to 2017, she worked as a Financial Auditor at Deloitte. From 2017 to 2018 she worked as a Legislation and Compliance Speacialist in subsidiary of Garanti BBVA Group. Since 2018, she is working as a Financial Crimes Investigation and Compliance Speacialist at Akbank Compliance Department.
\end{IEEEbiography}

\EOD

\end{document}